 \documentclass[timesreview， 10pt]{elsarticle}

\usepackage{amssymb}
\usepackage{amsmath}
\usepackage{multirow}
\usepackage{booktabs}
\usepackage{xcolor}

\journal{}

\begin{document}

\begin{frontmatter}



\title{SAGS: Self-Adaptive Alias-Free Gaussian Splatting for Dynamic Surgical Endoscopic Reconstruction}


\author[uts,siat]{Wenfeng Huang}
\author[siat]{Xiangyun Liao}
\author[siat]{Yinling Qian}
\author[sia]{Hao Liu}
\author[sia]{Yongming Yang}
\author[uts]{Wenjing Jia}
\author[siat]{Qiong Wang\corref{cor2}}

\address[uts]{University of Technology Sydney, Australia}
\address[siat]{Shenzhen Institute of Advanced Technology, Chinese Academy of Sciences, China}
\address[sia]{Shenyang Institute of Automation, Chinese Academy of Sciences, China}

\cortext[cor2]{Corresponding author: Yinling Qian and Qiong Wang (Email: wangqiong@siat.ac.cn)}

\begin{abstract}
Surgical reconstruction of dynamic tissues from endoscopic videos is a crucial technology in robot-assisted surgery. The development of Neural Radiance Fields (NeRFs) has greatly advanced deformable tissue reconstruction, achieving high-quality results from video and image sequences. 
However, reconstructing deformable endoscopic scenes remains challenging due to aliasing and artifacts caused by tissue movement, which can significantly degrade visualization quality. 
The introduction of 3D Gaussian Splatting (3DGS) has improved reconstruction efficiency by enabling a faster rendering pipeline. 
Nevertheless, existing 3DGS methods often prioritize rendering speed while neglecting these critical issues. 
To address these challenges, we propose SAGS, a self-adaptive alias-free Gaussian splatting framework. We introduce an attention-driven, dynamically weighted 4D deformation decoder, leveraging 3D smoothing filters and 2D Mip filters to mitigate artifacts in deformable tissue reconstruction and better capture the fine details of tissue movement. Experimental results on two public benchmarks, EndoNeRF and SCARED, demonstrate that our method achieves superior performance in all metrics of PSNR, SSIM, and LPIPS compared to the state of the art while also delivering better visualization quality.
\end{abstract}


\end{frontmatter}



\section{Introduction} \label{sec:intro}

3D reconstruction of deformable tissue structures from dynamic endoscopic videos represents an essential cornerstone in modern robotic-assisted surgical interventions, significantly enhancing navigation, precision, and overall patient outcomes~\cite{zha2023endosurf}. High-quality, real-time 3D reconstructions facilitate numerous intraoperative clinical applications, including augmented reality (AR)-based visualization, robotic surgery automation, immersive surgical training, and precise surgical planning~\cite{liu2024endogaussian, liu2024endogaussian2, endogs}.

Early developments in the medical scene reconstruction primarily relied on conventional depth estimation techniques and simultaneous localization and mapping (SLAM)-based frameworks~\cite{song2017dynamic}. Classic methods such as E-DSSR~\cite{long2021dssr} and Surfelwarp~\cite{gao2019surfelwarp} established preliminary successes by integrating stereo depth cues and dynamic surface tracking. However, these approaches struggle to accurately handle severe non-rigid tissue deformations, significant occlusion by surgical instruments, and complex dynamics typically encountered in real surgical environments~\cite{endogs,zha2023endosurf}. This fundamental limitation motivated further exploration into more robust and scalable representations.

The emergence of Neural Radiance Fields (NeRFs)~\cite{nerf} significantly shifted the paradigm toward implicit volumetric representations, achieving photorealistic quality in novel-view synthesis and continuous 3D scene modelling~\cite{xia2023survey}.  
NeRF leverages multi-layer perceptrons (MLPs) to implicitly encode volumetric densities and radiance, thereby attaining markedly higher visual fidelity than traditional discrete approaches.  
Dynamic extensions—such as Dynamic NeRF (D-NeRF)~\cite{pumarola2020d}, Neural Volumes~\cite{pumarola2020d}, and Temporal-Interpolation NeRF (TiNeuVox)~\cite{TiNeuVox}—further generalise this framework to temporally evolving scenes, while LerPlane~\cite{lerplane} reduces complexity by factorising the volume into a set of explicit planes, accelerating optimisation and improving near–real-time applicability. Within the medical domain, EndoNeRF~\cite{endonerf} represents the first attempt to apply NeRF to 3D surgical reconstruction. By integrating a static radiance field with a temporal deformation field, EndoNeRF can jointly encode geometry and temporal motion from a limited set of images, thereby enabling flexible 3D scene synthesis in dynamic surgical settings. EndoSurf~\cite{zha2023endosurf} embeds signed-distance fields within a radiance-field backbone to impose explicit geometric constraints, yielding smoother and more precise surfaces that are crucial for surgical visualisation.
However, NeRF‐style approaches are inherently limited by their requirement to sample numerous points along each viewing ray and to perform an MLP evaluation at every sample. This high computational multiplicity leads to long training cycles, substantial memory footprints, and rendering latencies that are incompatible with the real-time requirement of intra-operative surgical guidance, thereby motivating the exploration of explicit, real-time representations such as Gaussian Splatting~\cite{3dgs,xia2023survey}.  

Recent advancements introduced explicit representations via 3D Gaussian Splatting (3DGS), overcoming critical limitations of implicit models. 3DGS represents scenes explicitly with anisotropic Gaussian primitives optimized through differentiable rasterization, enabling rapid inference speeds and real-time rendering capabilities~\cite{3dgs,xia2023survey}. Groundbreaking studies like 3DGS demonstrated substantial performance improvements, achieving real-time frame rates while maintaining visual fidelity competitive with state-of-the-art NeRF-based methods~\cite{3dgs}. Extending 3DGS to dynamic scenes, 4D Gaussian Splatting (4DGS) integrates temporal deformation fields directly into Gaussian primitives, offering an efficient representation for dynamic scenes by using lightweight neural deformation networks to model Gaussian trajectories over time~\cite{4dgs}. This development significantly improves rendering speed and storage efficiency compared to previous methods, thus proving highly suitable for dynamic surgical scene reconstruction.

Motivated by these limitations, a growing body of work has adopted Gaussian-Splatting to accelerate the reconstruction of dynamic surgical scenes.    
EndoGS~\cite{endogs} boosts monocular performance by combining depth-guided spatio-temporal weighting with surface-aligned regularisation, thus alleviating severe occlusions.  
EndoGaussian~\cite{liu2024endogaussian, liu2024endogaussian2} introduces holistic Gaussian initialisation from depth estimation and incorporates a lightweight spatio-temporal tracker to cope with large deformations.
Although GS methods can achieve real-time rendering, endoscopic surgery still presents unresolved challenges: large non-rigid tissue motion, instrument-induced occlusions, and an uneven or sparse Gaussian distribution often give rise to aliasing artefacts and inaccurate geometry. Existing GS frameworks primarily optimise for speed and do not explicitly address alias suppression or deformation robustness.  

To overcome these shortcomings, we introduce \textbf{SAGS}, an attention-driven, alias-free Gaussian-splatting framework specifically designed for dynamic endoscopic reconstruction. SAGS suppresses high-frequency artefacts while employing a self-adaptive deformation decoder to capture complex tissue motion, thereby delivering high-fidelity 3D reconstructions suitable for the reconstruction of deformable endoscopic tissues. %The contributions of this paper include:\\

\begin{figure*}[ht]
	\centering
	\includegraphics[width = 1\linewidth]{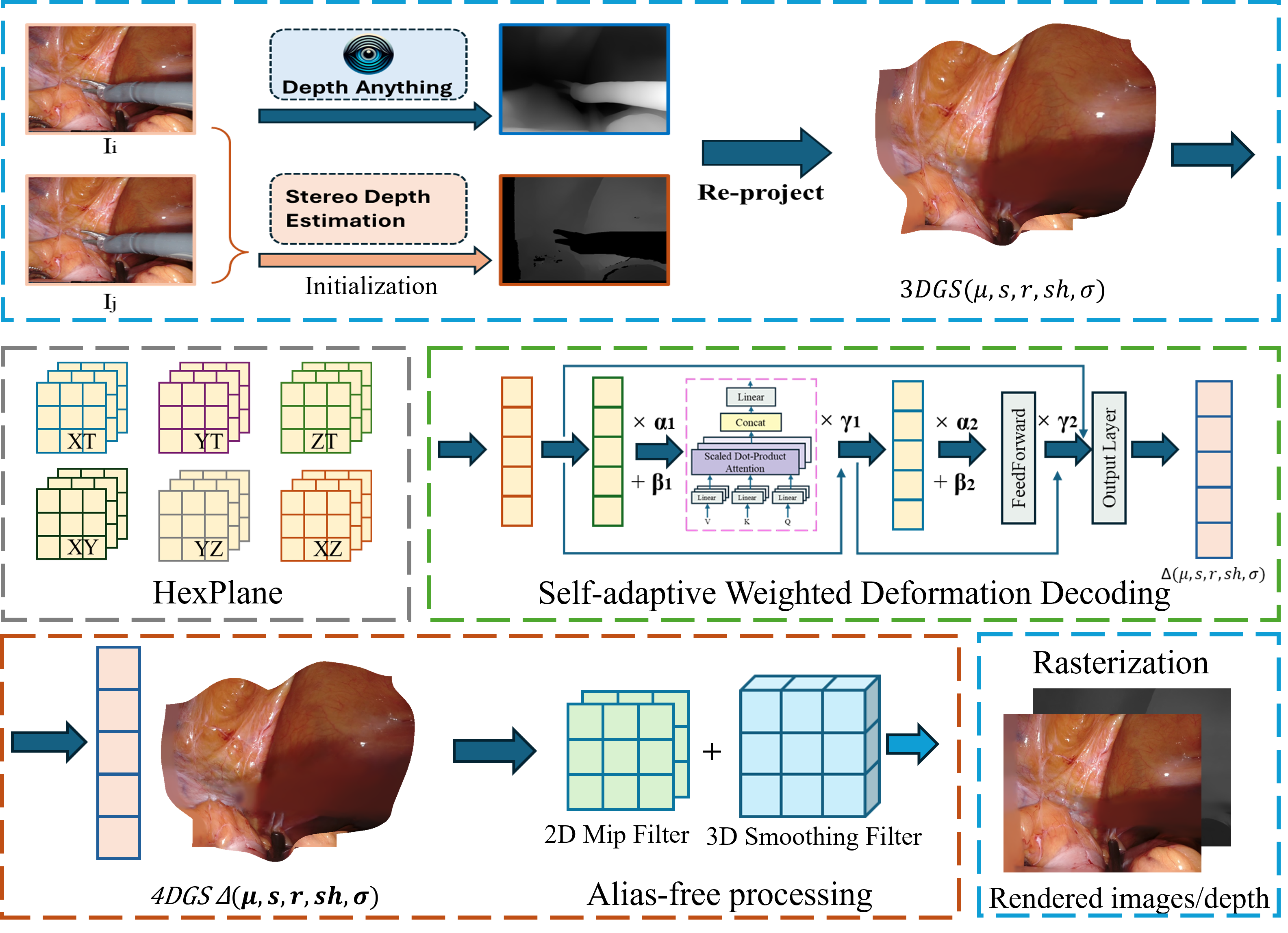}
 \vspace{-2em}
 \label{overall}
\caption{The overall pipeline of the proposed SAGS framework. Depth maps from monocular and stereo estimation are re-projected to initialize 3D Gaussians, which are then refined using HexPlane encoding and a self-adaptive weighted deformation decoder MLP for deformation modeling. Alias-free filtering with Mip and smoothing filters is subsequently applied, and finally, rasterization generates high-fidelity rendered images and depth.}

	\label{SAGS}
 % \vspace{-1em}
\end{figure*}

The contributions of this paper include:\\
\textbf{(1)} We propose a self-adaptive weighted deformation decoder, a multi-head attention-based mechanism capable of dynamically weighting Gaussian attributes, significantly enhancing the ability to model deformations in complex endoscopic scenes.\\
\textbf{(2)} We employ 3D smoothing filters and 2D Mip filters to achieve alias-free processing, effectively reducing artifacts in the reconstruction of deformable tissues.\\
\textbf{(3)} Experimental results on the public benchmarks EndoNeRF~\cite{endonerf} and SCARED~\cite{scared} show that our method performs better than state-of-the-art approaches in PSNR, SSIM, and LPIPS metrics while also delivering enhanced visual reconstruction quality.

\section{Related Works}

\subsection{3D Reconstruction}

3D reconstruction has experienced rapid development~\cite{pr2}, driven by various methods and applications ranging from traditional computer vision techniques to advanced neural rendering frameworks. Early classical methods such as Structure-from-Motion (SfM)~\cite{westoby2012structure},  and traditional volumetric rendering have laid fundamental groundwork for subsequent innovations~\cite{chen2019point,yao2020blendedmvs}. Real-time non-rigid capture emerged with DynamicFusion (2015)~\cite{newcombe2015dynamicfusion}, which fused depth maps while estimating dense deformation fields. Li et al.~\cite{pr6} proposed a weighted 3D volume reconstruction method from slice data based on a modified Allen–Cahn equation.
Shi et al.~\cite{pr3} introduced an edge-guided framework that improves 3D reconstruction from RGB images and sketches.

More recently, neural implicit representations, particularly NeRF~\cite{nerf}, have significantly advanced the quality and realism of scene reconstructions by representing 3D scenes implicitly through continuous volumetric functions optimized via differentiable rendering techniques~\cite{pumarola2020d}. And more recent developments, such as D-NeRF~\cite{pumarola2020d} and NeRFies~\cite{park2021nerfies}, integrated deformation fields to capture dynamic scenes effectively. To tackle the computational complexity and slow inference speed inherent to NeRF, several follow-up methods emerged. Instant neural graphics primitives employ a multi-resolution hash grid to drastically speed up both the optimization and rendering phases of Neural Radiance Field models~\cite{muller2022instant, hong2022headnerf}, while explicit representations, such as Plenoxels, provide real-time rendering capabilities without relying on neural networks~\cite{fridovich2022plenoxels}. H\textsubscript{2}O-NeRF reconstructs radiance fields for two-hand-held objects by combining SDF-based semantic cues with view-dependent visibility masks, improving completeness and view consistency under severe hand occlusion~\cite{tvcg}. MS-NeRF enhances NeRF performance in reflective and refractive scenes by introducing parallel sub-space radiance fields, achieving higher PSNR with minimal computational overhead~\cite{yin2025ms}. STGC-NeRF enforces spatial-temporal geometric consistency for dynamic LiDAR scenes, improving reconstruction accuracy from sparse, low-frequency input data~\cite{yu2025stgcnerf}. However, most of the NeRF-based methods remain burdened by substantial computational demands and a substantial memory footprint.

Recently, 3DGS has emerged as a promising technique, explicitly representing scenes with optimized anisotropic Gaussian primitives. This method enabled rapid rendering and improved reconstruction quality by directly optimizing Gaussian attributes from scene images~\cite{3dgs}. Subsequent enhancements to 3DGS further improved performance, including 3DGS for anti-aliased rendering via multi-scale design~\cite{yan2024multi}, and Mip-Splatting for alias-free representation~\cite{yu2024mip}. SpotLessSplats improves the robustness of 3D Gaussian Splatting by suppressing transient distractors using pre-trained features and robust optimisation, enabling high-quality reconstruction in casual capture settings~\cite{sabour2025spotlesssplats}. EDGS improves efficiency in dynamic Gaussian Splatting by modeling sparse, time-variant attributes, significantly reducing redundancy and accelerating rendering for monocular videos~\cite{kwon2025instruct4dgs}. Extending these concepts into 4D scenarios, 4DGS was proposed to handle temporal variations explicitly, providing the capabilities that can render dynamic scenes in real-time~\cite{4dgs}. Instruct-4DGS enables efficient dynamic scene editing by separating static and dynamic components within a 4D Gaussian framework, significantly reducing editing time while preserving visual fidelity~\cite{kwon2025instruct4dgs}. Overall, these methods collectively pushed the boundaries of 3D and 4D reconstruction, significantly enhancing realism, computational efficiency, and adaptability to dynamic scenes. Jiang et al~\cite{pr4}. proposed a real-time point-splatting framework for dynamic hand reconstruction with photorealistic rendering.

\subsection{3D Reconstruction for Medical Applications}

3D reconstruction techniques have been increasingly important in medical scenarios, particularly for dynamic surgical environments. Traditional reconstruction approaches such as SfM~\cite{westoby2012structure} and SLAM-based methods~\cite{chen2018slam, zhou2019real} have facilitated initial explorations in reconstructing medical scenes. SfM reconstructs 3D models by estimating camera poses and sparse point clouds from image collections, but it typically struggles with dynamic and textureless scenes common in endoscopic procedures. In contrast, SLAM-based methods integrate camera localization and dense mapping simultaneously, offering more robust performance in minimally invasive surgery~\cite{chen2018slam, zhou2019real}. Machucho-Cadena et al.~\cite{pr5} applied geometric algebra methods for ultrasound probe tracking, tumor segmentation, and 3D reconstruction in medical imaging. Zhang et al.~\cite{pr2} proposed a domain-adaptive method for 3D microvascular reconstruction from OCT angiography images. 

Recent developments in neural rendering, especially NeRF~\cite{nerf}, have sparked significant interest in reconstructing dynamic surgical scenes. Methods such as EndoNeRF~\cite{endonerf} and D-NeRF~\cite{pumarola2020d} have effectively modeled dynamic tissues by training neural fields for deformation and canonical density. EndoNeRF specifically adapts NeRF to endoscopic contexts, integrating dynamic deformation fields with neural radiance for realistic scene reconstruction~\cite{corona2022mednerf, psychogyios2023realistic}. To improve rendering speed and training efficiency, LerPlane~\cite{lerplane} utilizes 4D representations by extending 3D spaces with temporal dimensions, significantly accelerating the reconstruction process. However, NeRF-based pipelines are fundamentally constrained by lengthy training cycles, slow inference, and considerable memory footprints, which collectively hinder their practical deployment in the operating theatre.

As an explicit representation approach, 3DGS has recently emerged, demonstrating impressive real-time rendering capabilities and high-quality reconstructions through anisotropic Gaussians~\cite{3dgs}. Adaptations of 3DGS for dynamic scenes, such as  EndoGS~\cite{endogs} and EndoGaussian~\cite{liu2024endogaussian, liu2024endogaussian2}, have achieved superior reconstruction quality by addressing deformation tracking and spatial-temporal coherence. 
EndoGS employs surface-aligned regularization to reduce artifacts and enhance surface consistency, proving robust against occlusions common in surgical procedures~\cite{endogs}. 
Similarly, EndoGaussian introduces holistic Gaussian initialization and spatio-temporal tracking for effective real-time performance~\cite{liu2024endogaussian, liu2024endogaussian2}. HFGS specifically targets high-frequency reconstruction issues, enhancing both spatial and temporal fidelity in endoscopic videos~\cite{Zhao2024}. Other notable methods such as Deform3DGS~\cite{yang2024deform3dgs} integrate deformation fields and surface alignment into the 3D Gaussian framework to enhance reconstruction accuracy and surface details. Despite these advancements, existing methods still face challenges related to artifact reduction, spatial-temporal coherence, and computational efficiency, motivating continuous development in the field. Nevertheless, most existing GS-based pipelines are still optimised primarily for speed and therefore devote limited capacity to learning alias-suppression mechanisms and fine-grained deformation cues; consequently, specular ringing, texture drift, and subtle folding artefacts remain visible in challenging frames, revealing a persistent gap in alias-aware, deformation-adaptive modelling that motivates the proposed \emph{SAGS} framework.

\section{Methodology}

Endoscopic surgical scenes are characterised by rapid, non-rigid tissue motion, severe occlusions from instruments, and highly specular, spatially varying illumination.  
These factors impose stringent requirements on any 3D representation used for intra-operative guidance: it must preserve fine geometric detail without introducing aliasing artefacts, and remain robust to large, topology-changing deformations.  
Traditional NeRF-based methods struggle to satisfy these constraints. And secondly,  their implicit volumetric formulation incurs substantial computational latency, whereas classical SLAM pipelines fail to model the continuous tissue motion observed in laparoscopy.  
Consequently, there is a growing interest in explicit, point-based encodings that can be updated and rendered faster while still providing photorealistic quality.

\subsection{Preliminary of 3D Gaussian Splatting}

3DGS~\cite{3dgs} provides fast rendering capabilities and superior 3D representation performance. It represents scenes explicitly through point clouds, which models each point cloud as a 3DGS characterized by a center point $\mu$ (\textit{a.k.a.}, the mean), as well as a covariance matrix $\Sigma$  as:
\begin{equation}
    G(x) = e^{-\frac{1}{2}(x-\mu)^T \Sigma^{-1}(x-\mu)}. 
% \tag{1}
\end{equation}

% To render 3D Gaussians in 2D space, they are projected and transformed into the 2D plane using the covariance matrix $\Sigma'$, computed as $\Sigma' = J W \Sigma W^T J^T$, where $\Sigma'$ denotes the 2D covariance matrix, $W$ represents the view transformation, and $J$ is the affine approximation of the projective process. 
When projecting 3-D Gaussians onto the image plane, each Gaussian’s covariance is transformed into a 2D covariance matrix via
\[
\Sigma' \;=\; J\,W\,\Sigma\,W^{T}J^{T},
\]
where \(\Sigma\) is the original 3D covariance, \(W\) represents the view-dependent rigid transformation, and \(J\) is the Jacobian matrix of the projection operation.

The covariance matrix $\Sigma$ is expressed as: $\Sigma = R S S^T R^T$, where $R$ defines the rotation, and $S$ specifies the scale, to ensure positive semi-definiteness. Rendering pixel colors $C(p)$ is achieved through point-based volume rendering, which combines color contributions and opacities of Gaussians along the ray:
\begin{equation}
    C(p) = \sum_{i \in N} c_i \alpha_i \prod_{j=1}^{i-1} (1 - \alpha_j). 
\end{equation}
Here, the opacity $\alpha_i$ for each Gaussian is computed as:
\begin{equation}
    \alpha_i = \sigma_i e^{-\frac{1}{2} (p-\mu_i)^T \Sigma'^{-1} (p-\mu_i)}. 
\end{equation} 
In this framework, $\mu_i$ specifies the Gaussian's position, $c_i$ represents its color, and $\sigma_i$ indicates its opacity. 
To account for view-dependent effects, spherical harmonics are employed for color modeling. Each explicit 3D Gaussian is modeled by a set of characteristics: its position $\mu \in \mathbb{R}^3$, scaling factor $s \in \mathbb{R}^3$, rotation factor $r \in \mathbb{R}^4$, spherical harmonic (SH) coefficients $sh \in \mathbb{R}^k$ (where $k$ denotes the number of SH functions), and opacity $\sigma \in \mathbb{R}$. Collectively, these attributes define the Gaussian as $(\mu, s, r, sh, \sigma)$.

\subsection{Point Cloud Information Acquisition}

Accurate depth cues are indispensable for dynamic endoscopic reconstruction, where narrow baselines, specular highlights, and rapid non-rigid tissue motion make reliable correspondence estimation particularly challenging.  In such confined surgical scenes, point clouds must therefore be initialised from either monocular or binocular cues before they can be refined by our Gaussian-splatting pipeline.\\
\textbf{Monocular Depth:}
\label{sec:mono}
Recent advances in large-scale depth pre-training have markedly improved single-frame depth estimation, even under the extreme lighting and texture conditions of minimally invasive surgery.  
Inspired by previous work~\cite{liu2024endogaussian, liu2024endogaussian2}, we adopt \emph{Depth Anything}~\cite{depthanything}, which is optimised on billions of images with synthetic and sparsely supervised depth, and has demonstrated strong zero-shot generalisation to laparoscopic footage.   
Given an endoscopic image \(I_{i}\) from time step \(T\), the network predicts a dense depth map:
\[
  D_{i} = \text{DepthAnything}(I_{i}),
\]  
which we then back-project through the intrinsic calibration to generate an initial partial point cloud \(P_{i}\) following the reprojection scheme in EndoNeRF~\cite{endonerf}.
\\
\textbf{Binocular Depth:}
Building on the previous work~\cite{liu2024endogaussian, liu2024endogaussian2}, binocular depth estimation is achieved by using adjacent stereo inputs $I_{i}$ and $I_{j}$ to compute the depth of the remaining %left 
frame $D_{i}$ using stereo depth prediction methods~\cite{depth}. The depth $D_{i}$ is then processed through the same re-projection pipeline described in Monocular Depth to generate the corresponding point cloud $P_{i}$.

\subsection{4D Representation for Deformable Tissue Reconstruction}
In dynamic endoscopic surgery, tissue motion is continuous and highly non-rigid, necessitating %demanding 
a 4D representation that can evolve over time. While standard 3DGS captures static geometry efficiently, it lacks the temporal expressiveness needed for live surgical scenes. 4DGS~\cite{4dgs} tackles this gap by proposing a deformable time-series representation in which each Gaussian primitive can change its position, shape, and appearance across frames.

The 4D representation in the proposed SAGS models the Gaussian deformations of deformable tissues. This involves not only learning the Gaussian attributes—position $\mu \in \mathbb{R}^3$, scaling factor $s \in \mathbb{R}^3$, rotation factor $r \in \mathbb{R}^4$, spherical harmonic (SH) coefficients $sh \in \mathbb{R}^k$, and opacity $\sigma \in \mathbb{R}$—but also tracking their deformations, which are defined as a set $\Delta GS$. To drive the deformation fields, we employ the HexPlane~\cite{cao2023hexplane} with a resolution of $D_1$ and $D_2$. The HexPlane comprises six planes: $P_{XY}, P_{XZ}, P_{YZ}, P_{XT}, P_{YT},$ and $P_{ZT}$, where the first three are spatial planes and the latter three are spatiotemporal planes. 

The HexPlane encodes Gaussian information $I$, where $I \in \mathbb{R}^{h \times D_1 \times D_2}$ and $h$ represents the hidden space. The encoded voxel information $I_{\text{voxel}}$ for point $(\mu, t)$ can be extracted as:
\begin{align}
    I_{\text{voxel}}(\mu, t) = 
    &\mathcal{F}(I_{XY}, x, y) \odot \mathcal{F}(I_{XZ}, x, z) \odot \mathcal{F}(I_{YZ}, y, z) \nonumber \\
    &\odot \mathcal{F}(I_{XT}, x, t) \odot \mathcal{B}(I_{YT}, y, t) \odot \mathcal{F}(I_{ZT}, z, t).
\end{align}
Here, $\mathcal{F}$ denotes the bilinear interpolation operation used to obtain the nearest voxel Gaussian information, and $\odot$ represents the element-wise multiplication. This voxel encoding mechanism ensures the integration of spatial and temporal features, which are critical for accurately reconstructing the deformation fields. However, the aforementioned 4DGS variant that relies solely on the HexPlane representation is unable to capture the complex, non-rigid deformations present in dynamic surgical scenes. To overcome this problem, we proposed a \emph{Self-adaptive Weighted Deformation Decoder}.

% \subsection{4D Deformation Decoder}
\subsection{Self-adaptive Weighted Deformation Decoding}
Endoscopic scene reconstruction is uniquely challenging: tissues undergo large, non-rigid deformations, surgical tools create severe and view-dependent occlusions, lighting is highly specular and spatially varying, and camera motion is restricted to narrow baselines.  
These factors hinder stable correspondence estimation and make it difficult for conventional neural encoders to predict temporally coherent geometry and appearance.  
To address these issues, and inspired by recent advances of MLPs~\cite{touvron2022resmlp, liu2021pay}, we introduce a dynamically weighted multi-head self-attention module integrated with a deformation-aware decoder, named the self-adaptive deformation decoder, to model and decode the Gaussian attribute deformations. Unlike traditional methods that rely solely on fixed-weight MLPs, our approach introduces learnable dynamic weights to adaptively focus on spatial-temporal features, enabling more accurate and robust deformation predictions for each attribute.

% \subsubsection{Multi-head Self-Attention}
The dynamic weight mechanism assigns adaptive importance to different features during the self-attention computation. 
A fixed combination of attention and MLP outputs would fail to account for the varying demands of endoscopic scenes, where large-scale motions demand global coherence while fine-scale tissue variations require local refinement.
Specifically, for each Gaussian attribute deformation, the dynamic weights $\gamma_1$ and $\gamma_2$ adjust the contributions of the self-attention output and the MLP output, respectively. These weights are learnable and initialized with small values to allow gradual learning during training. 
The self-attention approach can be defined as:
\begin{equation}
\text{Attention}(Q, K, V) = \text{softmax}\left(\frac{QK^\top}{\sqrt{d_k}}\right)V.
\end{equation}
Here, \(Q\) (Qurey), \(K\) (Key), and \(V\) (Value) are matrices obtained from the encoded voxel features \(I_{\mathrm{voxel}}(\mu, t)\), and each key vector has dimensionality \(d_{k}\). The dimensionality of the key vectors is $d_k$. The attention is applied across multiple heads: $\text{MSA}(Q, K, V) = \text{Concat}(\text{head}_1, \text{head}_2, \ldots, \text{head}_h)W^O$, where $\text{head}_i$ donates the output of the $i$-th attention head, and $W^O$ is a learnable projection matrix.
% \subsubsection{Dynamic Weighted Feature Aggregation}

% To adaptively combine the self-attention and MLP outputs, we introduce a dynamically weighted decoding mechanism, defining $y$ as the learnable deformable attributes:  

% As shown on Fig.~\ref{overall}, to adaptively combine the self-attention and MLP outputs, we also adapt the dynamically weighted mechanism, where the learnable deformable attributes, denoted as $y$ is are computed as:

As shown in Fig.~\ref{overall}, we employ a dynamically weighted mechanism to adaptively combine the outputs of the self-attention and MLP branches. 
In this design, the learnable deformable attributes, denoted as $y$, are obtained by aggregating the contributions from both pathways under dynamic weights (\textit{i.e.}, $\alpha_1$, $\beta_1$, $\alpha_2$, $\beta_2$, $\gamma_1$, and $\gamma_2$). This mechanism allows the model to flexibly balance global consistency captured by self-attention and local detail refinement provided by MLPs, ensuring that the deformation representation adapts effectively to varying tissue motions. 

This self-adaptive decoding is formulated as:
\begin{equation}
    y = \text{Affine}_\text{post}(y') + \gamma_2 \cdot \text{MLP}(\text{Affine}_\text{post}(y')) + x,
\end{equation}
where
\begin{equation}
    y' = \text{Affine}_\text{pre}(x) + \gamma_1 \cdot \text{MSA}( \text{Affine}_\text{pre}(x)).
\end{equation}

% Here, $\text{Affine}(x)$ applies an affine transformation: $\text{Affine}(x) = \alpha \cdot x + \beta$  to the input $x$,  $\text{MSA}(x)$ represents the multi-head self-attention module and $\text{MLP}(x)$ denotes a feed-forward network applied to the input features. 
%
Here, $\text{Affine}_\text{pre}(x)$ and $\text{Affine}_\text{post}(x)$ denote two learnable Affine transformations applied to the input features, formulated as $\text{Affine}(x) = \alpha \cdot x + \beta$, where $\alpha$ and $\beta$ represent the scaling and shifting parameters, respectively. 
The $\text{Affine}_\text{pre}$ transformation serves to normalize and re-scale the features before they enter the attention block, while $\text{Affine}_\text{post}$ adjusts the outputs after feature aggregation, ensuring stable training and effective feature fusion.
The two-stage pre-post Affine transformation stabilizes feature scaling and shifting before and after each branch, which regularizes feature magnitudes and facilitates residual learning. 
The weighted residual formulation, governed by $\gamma_1$ and $\gamma_2$, provides a controllable trade-off between global and local cues, rather than enforcing a rigid fusion. 
This operation serves as a lightweight linear adaptation layer, ensuring that the input features are properly normalized and rescaled before entering the self-attention and MLP branches. 
Learning the affine parameters $\alpha_1$, $\alpha_2$, $\beta_1$, and $\beta_2$ further increases the flexibility of the decoder by allowing feature-level adjustments that facilitate stable training and improve the expressiveness of deformation modeling.

This design leverages the complementary strengths of the two modules: self-attention is particularly effective at capturing long-range dependencies and preserving global geometric consistency, whereas the MLP component is more adept at modeling local nonlinear variations and fine-grained tissue deformations. 
Following the previous work~\cite{liu2024endogaussian, liu2024endogaussian2}, we use four small MLPs in the output layers.
Instead of assigning fixed contributions, the dynamically learnable weights (\textit{i.e.}, $\alpha_1$, $\alpha_2$, $\beta_1$, $\beta_2$, $\gamma_1$, and $\gamma_2$) regulate the relative influence of these two information pathways, enabling the network to emphasize global coherence under substantial movements and viewpoint shifts, while simultaneously prioritizing localized corrections when detailed tissue structures undergo fine-scale motion.  

Within the Self-adaptive Weighted Deformation Decoding module, the residual outputs from the attention and MLP branches are not only combined through dynamically learned weights but are also propagated to update the Gaussian parameters directly. 
In this way, the adaptive weighting directly governs how global coherence and local deformation cues translate into geometry refinement. 

%The updated Gaussian attributes are thus obtained by applying the learned $\Delta$ to their initial values, yielding:
As the final step of the self-adaptive weighted deformation decoding, the residuals $\Delta$ derived from the attention–MLP aggregation are applied to the initial Gaussian parameters. 
In this way, the Gaussian attributes are iteratively refined and updated as:
\begin{align}
    \mu' &= \mu + \Delta\mu, \quad s' = s + \Delta s, \quad r' = r + \Delta r, \\
    sh' &= sh + \Delta sh, \quad \sigma' = \sigma + \Delta\sigma.
\end{align}
Here, $\mu \in \mathbb{R}^3$ denotes the Gaussian mean, representing the 3D spatial position of the primitive.  
$s \in \mathbb{R}^3$ encodes the anisotropic scaling factors, which control the spatial extent of the Gaussian along the three principal axes.  
$r \in \mathbb{R}^4$ corresponds to the quaternion rotation, defining the orientation of the Gaussian ellipsoid in 3D space.  
$sh \in \mathbb{R}^k$ represents the spherical harmonic coefficients that model view-dependent color variations, enabling photorealistic appearance representation.  
Finally, $\sigma \in \mathbb{R}$ denotes the opacity term, governing the transparency and blending behavior of the Gaussian during rasterization.  

This novel dynamic attention-based deformation decoder significantly enhances the representation and reconstruction quality for deformable tissues by combining attention-based global feature aggregation and local feature refinement, allowing the model to effectively capture fine-grained details and large-scale deformations in a highly adaptive manner.

\subsection{Alias-Free Processing}

Surgical dynamic reconstruction is particularly prone to visual artefacts: specular highlights generated by the endoscope light, narrow stereo baselines, and rapid non-rigid tissue motion combine to produce strong aliasing, ringing, and frame-to-frame flicker.  
In our 4D reconstruction framework, these aliasing and high-frequency artefacts pose a major obstacle to both quantitative fidelity and clinical usability.  
To mitigate them, we integrate 3D smoothing filters together with 2D Mip filters, drawing on the anti-aliasing principles of Mip-Splatting~\cite{Yu2024MipSplatting,mip2}.  
Acting in tandem, the volumetric (3D) and image-space (2D) filters attenuate high-frequency noise during Gaussian optimisation and during the final projection step, respectively, thereby yielding temporally consistent, artefact-free reconstructions.

The 3D smoothing filter~\cite{Yu2024MipSplatting} is applied to each Gaussian primitive to constrain its high-frequency components according to the Nyquist sampling theorem. Given the maximal sampling rate $\hat{v}_k$ of a Gaussian primitive $k$, we apply a filter for Gaussian low-pass $G_{\text{low}}$ with variance $\Sigma_{\text{low}}$ to regularize the Gaussian, defined as: $G_k(\mathbf{x})_{\text{reg}} = (G_k \ast G_{\text{low}})(\mathbf{x})$, where $\ast$ denotes the convolution operation. The convolution of two Gaussians results in another Gaussian, with the new covariance matrix given by $\Sigma_k + \Sigma_{\text{low}}$. The regularized Gaussian can be expressed as:
\begin{equation}
\mathcal{G}_k(\mathrm{x})_{\mathrm{reg}}=\sqrt{\frac{\left|\boldsymbol{\Sigma}_k\right|}{\left|\boldsymbol{\Sigma}_k+\frac{s}{\hat{\boldsymbol{\nu}_k}} \cdot \mathbf{I}\right|}} e^{-\frac{1}{2}\left(\mathbf{x}-\mathbf{p}_k\right)^T\left(\boldsymbol{\Sigma}_k+\frac{s}{\hat{\nu}_k} \cdot \mathbf{I}\right)^{-1}\left(\mathbf{x}-\mathbf{p}_k\right)},
\end{equation}
where $s$ is a scalar hyperparameter controlling the filter size, $\hat{v}_k$ is the maximal sampling rate for primitive $k$, and $\mathbf{p}_k$ represents the center of the Gaussian. By applying this filter, high-frequency artifacts in the volumetric domain are effectively reduced.

While 3D smoothing filters suppress high-frequency artifacts in the volumetric representation, aliasing can still occur during these Gaussians onto the 2D image plane. To handle this, we use 2D Mip filters~\cite{Yu2024MipSplatting} to approximate a box filter for each pixel in screen space, defined as:
\begin{equation}
\mathcal{G}_k^{2 D}(\mathbf{x})_{\text {mip }}=\sqrt{\frac{\left|\Sigma_k^{2 D}\right|}{\left|\Sigma_k^{2 D}+s \mathbf{I}\right|}} e^{-\frac{1}{2}\left(\mathbf{x}-\mathbf{p}_k\right)^T\left(\Sigma_k^{2 D}+s \mathbf{I}\right)^{-1}\left(\mathbf{x}-\mathbf{p}_k\right)},
\end{equation}
where $\Sigma_k^{\text{2D}}$ represents the Gaussian’s covariance projected into 2D screen space, and $s$ is chosen to cover a single pixel. This filter mimics the integration of photons over a pixel’s area, ensuring the alignment of the rendered Gaussians and the pixel resolution, significantly reducing aliasing during zoom-out views or varying camera distances.

By integrating 3D smoothing and 2D Mip filters, we ensure that the voxel features $I_{\text{voxel}}(\mu, t)$ in our 4D endoscopic reconstruction pipeline remain robust to both spatial and temporal aliasing. The smoothed and filtered voxel information can be computed as:
\begin{align}
    I'_{\text{voxel}}(\mu, t) = 
    &\mathcal{F}(G_{XY}', x, y) \odot \mathcal{F}(G_{XZ}', x, z) \odot \mathcal{F}(G_{YZ}', y, z) \nonumber \\
    &\odot \mathcal{F}(G_{XT}', x, t) \odot \mathcal{F}(G_{YT}', y, t) \odot \mathcal{F}(G_{ZT}', z, t),
\end{align}
where $G_{XY}', G_{XZ}', \dots$ represent Gaussians after 3D smoothing and 2D Mip filtering. This unified approach effectively suppresses artifacts in the 4D deformation fields, ensuring visually coherent and high-fidelity endoscopic scene reconstruction.

Following the previous work~\cite{liu2024endogaussian,liu2024endogaussian2}, we adopt similar loss functions to optimize our framework, combining rendering constraints and spatio-temporal smoothness constraints into a unified objective: $L = \lambda_1 L_\text{color} + \lambda_2 L_\text{depth} + \lambda_3 L_\text{spatial} + \lambda_4 L_\text{temporal}$, where $\lambda_{1,2,3,4}$ are balancing weights for color rendering, depth consistency, spatial regularization, and temporal smoothness, respectively.

% \section{Implementation Details}
\section{Experiments}

\subsection{Datasets}

Dynamic 3D reconstruction in invasive surgery must cope with centimetre-scale tissue deformations, strong specular reflections from the endoscope light, frequent occlusions by forceps or scissors, and a very limited camera baseline. To study these challenges systematically, we adopt two public benchmarks that have become standard in endoscopic reconstruction research.
In line with previous works~\cite{endonerf,zha2023endosurf,liu2024endogaussian, liu2024endogaussian2,endogs}, we trained and evaluated our method using two public benchmark datasets: EndoNeRF~\cite{endonerf} and SCARED~\cite{scared}.
\\\textbf{EndoNeRF~\cite{endonerf}}: This dataset was captured during in-vivo prostatectomy surgeries using stereo cameras. It includes two cases showcasing two distinct scenarios: tissue pulling and tissue cutting. The dataset presents significant challenges due to the irregular tissue deformation and the movement caused by surgical tools. \\ \textbf{SCARED~\cite{scared}}: The SCARED dataset was captured using a DaVinci endoscope and a projector, providing RGB-D data of porcine cadaver abdominal anatomies. The dataset provides seven training sequences and two hidden test sequences. We follow the previous work~\cite{liu2024endogaussian,liu2024endogaussian2,zha2023endosurf} to split the dataset for training and testing. 

\begin{table}[htpb]
\centering
\caption{Performance comparison on the EndoNeRF dataset with SOTAs.}
\resizebox{\textwidth}{!}{
\begin{tabular}{l l c c c}
\toprule
\textbf{Dataset} & \textbf{Method} & \textbf{PSNR↑} & \textbf{SSIM↑} & \textbf{LPIPS↓} \\ 
\midrule
\multirow{9}{*}{\textbf{EndoNeRF~\cite{endonerf}}}
& EndoNeRF~\cite{endonerf}                    & 36.062 & 0.933 & 0.089 \\
& EndoSurf~\cite{zha2023endosurf}             & 36.529 & 0.954 & 0.074 \\
& LerPlane-9k~\cite{lerplane}                 & 34.988 & 0.926 & 0.080 \\
& EndoGS~\cite{endogs}                        & 36.990 & 0.961 & 0.038 \\
& LerPlane-32k~\cite{lerplane}                & 37.384 & 0.950 & 0.047 \\ \cmidrule(l){2-5}
& EndoGaussian (Monocular)~\cite{liu2024endogaussian}      & 37.464 & 0.960 & 0.052 \\
& \textbf{SAGS (Monocular)}                        & \textbf{37.711} & \textbf{0.962} & \textbf{0.043} \\ \cmidrule(l){2-5}
& EndoGaussian (Binocular)~\cite{liu2024endogaussian} & 38.088 & 0.962 & 0.048 \\
& \textbf{SAGS (Binocular)}                  & \textbf{39.164} & \textbf{0.970} & \textbf{0.025} \\
\bottomrule
\end{tabular}
}
\label{comparison2}
\end{table}

\begin{table}[!h]
\centering
\caption{Performance comparison of SAGS with SOTA methods on EndoNeRF~\cite{endonerf} and SCARED~\cite{scared} datasets using binocular depths.}
\resizebox{\textwidth}{!}{
\begin{tabular}{lllccc}
\toprule
\textbf{Dataset} & \textbf{Task/Scene} & \textbf{Method} & \textbf{PSNR↑} & \textbf{SSIM↑} & \textbf{LPIPS↓} \\
\midrule
\multirow{8}{*}{\textbf{EndoNeRF~\cite{endonerf}}}
& \multirow{4}{*}{\textbf{Pulling}}
& EndoNeRF~\cite{endonerf} & 34.21 & 0.938 & 0.161 \\
& & EndoSurf~\cite{zha2023endosurf} & 35.00 & 0.956 & 0.120 \\
& & EndoGaussian~\cite{liu2024endogaussian} & 37.21 & 0.957 & 0.061 \\
& & \textbf{SAGS (Ours)} & \textbf{38.30} & \textbf{0.964} & \textbf{0.033} \\
\cmidrule(l){2-6}
& \multirow{4}{*}{\textbf{Cutting}}
& EndoNeRF~\cite{endonerf} & 34.19 & 0.932 & 0.151 \\
& & EndoSurf~\cite{zha2023endosurf} & 34.98 & 0.953 & 0.106 \\
& & EndoGaussian~\cite{liu2024endogaussian} & 38.44 & 0.968 & 0.043 \\
& & \textbf{SAGS (Ours)} & \textbf{39.51} & \textbf{0.972} & \textbf{0.022} \\
\midrule
\multirow{24}{*}{\textbf{SCARED~\cite{scared}}}
& \multirow{4}{*}{\textbf{d1k1}}
& EndoNeRF~\cite{endonerf} & 24.37 & 0.763 & 0.326 \\
& & EndoSurf~\cite{zha2023endosurf} & 24.40 & 0.769 & 0.319 \\
& & EndoGaussian~\cite{liu2024endogaussian} & 29.75 & 0.864 & 0.143 \\
& & \textbf{SAGS (Ours)} & \textbf{30.23} & \textbf{0.875} & \textbf{0.102} \\
\cmidrule(l){2-6}
& \multirow{4}{*}{\textbf{d2k1}}
& EndoNeRF~\cite{endonerf} & 25.73 & 0.828 & 0.240 \\
& & EndoSurf~\cite{zha2023endosurf} & 26.24 & 0.829 & 0.254 \\
& & EndoGaussian~\cite{liu2024endogaussian} & 30.90 & 0.871 & 0.189 \\
& & \textbf{SAGS (Ours)} & \textbf{33.53} & \textbf{0.915} & \textbf{0.070} \\
\cmidrule(l){2-6}
& \multirow{4}{*}{\textbf{d3k1}}
& EndoNeRF~\cite{endonerf} & 19.00 & 0.599 & 0.467 \\
& & EndoSurf~\cite{zha2023endosurf} & 20.04 & \textbf{0.649} & 0.441 \\
& & EndoGaussian~\cite{liu2024endogaussian} & 18.82 & 0.609 & 0.493 \\
& & \textbf{SAGS (Ours)} & \textbf{20.11} & 0.619 & \textbf{0.426} \\
\cmidrule(l){2-6}
& \multirow{4}{*}{\textbf{d6k1}}
& EndoNeRF~\cite{endonerf} & 24.04 & 0.833 & 0.464 \\
& & EndoSurf~\cite{zha2023endosurf} & 24.09 & 0.866 & 0.461 \\
& & EndoGaussian~\cite{liu2024endogaussian} & 25.69 & \textbf{0.871} & 0.372 \\
& & \textbf{SAGS (Ours)} & \textbf{25.73} & 0.856 & \textbf{0.304} \\
\cmidrule(l){2-6}
& \multirow{4}{*}{\textbf{d7k1}}
& EndoNeRF~\cite{endonerf} & 22.64 & 0.813 & 0.312 \\
& & EndoSurf~\cite{zha2023endosurf} & 23.42 & 0.861 & 0.282 \\
& & EndoGaussian~\cite{liu2024endogaussian} & 24.97 & 0.855 & 0.239 \\
& & \textbf{SAGS (Ours)} & \textbf{26.54} & \textbf{0.862} & \textbf{0.168} \\
\cmidrule(l){2-6}
& \multirow{4}{*}{\textbf{Average}}
& EndoNeRF~\cite{endonerf} & 26.31 & 0.815 & 0.303 \\
& & EndoSurf~\cite{zha2023endosurf} & 26.88 & 0.840 & 0.283 \\
& & EndoGaussian~\cite{liu2024endogaussian} & 29.40 & 0.857 & 0.220 \\
& & \textbf{SAGS (Ours)} & \textbf{30.56} & \textbf{0.866} & \textbf{0.161} \\
\bottomrule
\end{tabular}
\label{comparison}}
\end{table}

\subsection{Evaluation Metrics}
\begin{figure*}[t]
	\centering
	\includegraphics[width = 1\linewidth]{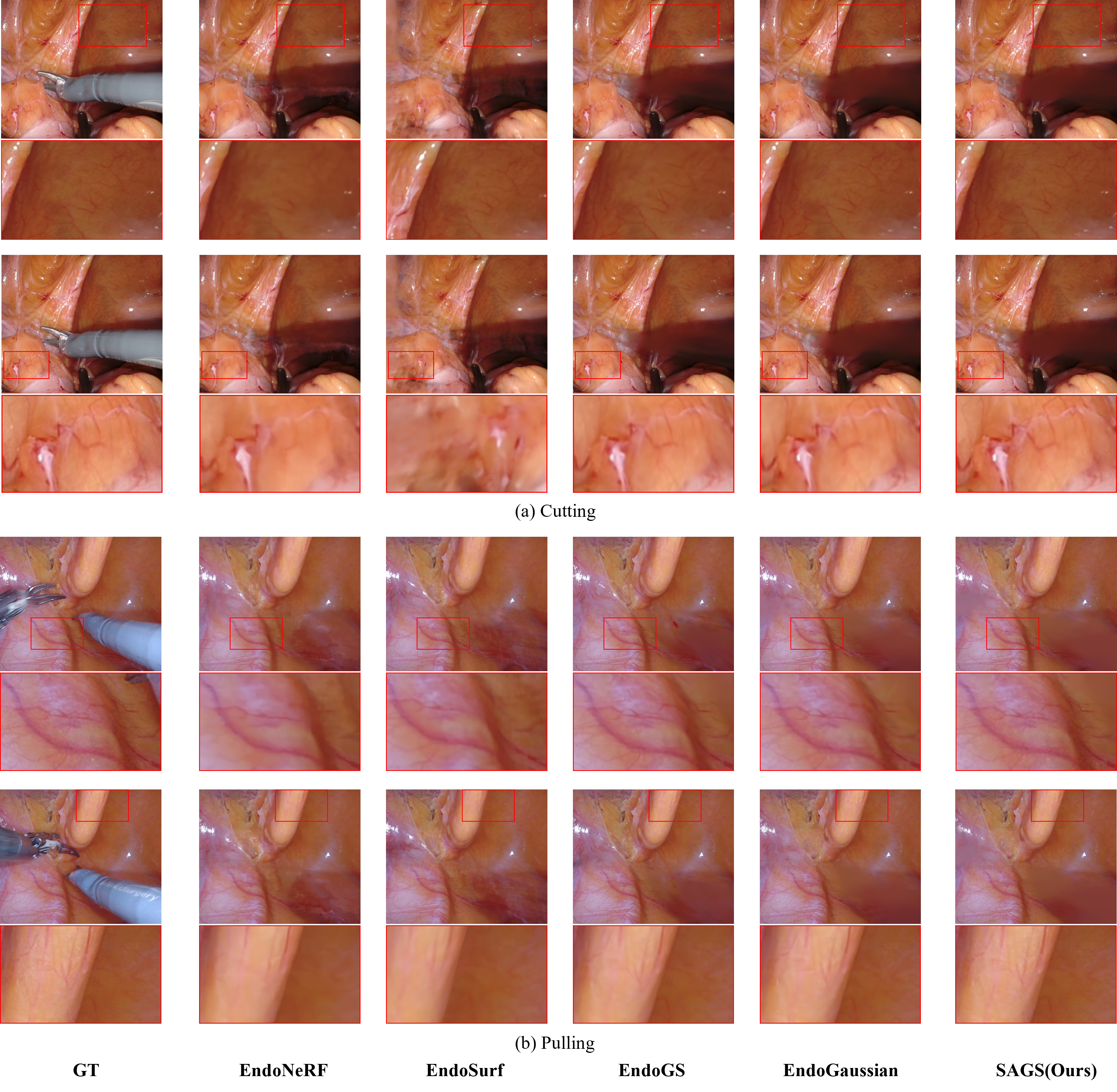}
 % \vspace{-2em}
        \caption{The qualitative result comparison between SOTAs and our proposed SAGS.}
\label{detail}
    % \vspace{-1.5em}
\end{figure*}
Precise visual feedback is vital in minimally invasive surgery, where even subtle geometric or textural errors can mislead a surgeon’s perception of tissue boundaries or tool–tissue interaction.  
To capture both pixel-level accuracy and perceptual plausibility in this high-risk setting, we employ three complementary metrics: Peak Signal-to-Noise Ratio (PSNR)~\cite{PSNR}, which quantifies reconstruction accuracy by measuring pixel-wise differences between the reconstructed and ground-truth images; Structural Similarity Index (SSIM)~\cite{SSIM}, which assesses perceived structural similarity by comparing luminance, contrast, and structural consistency; and Learned Perceptual Image Patch Similarity (LPIPS)~\cite{LPIPS}, which evaluates perceptual similarity using deep feature embeddings.

Following the previous work~\cite{liu2024endogaussian, liu2024endogaussian2}, we randomly sample 0.1\% of the points during the initialization stage to reduce redundancy and improve computational efficiency. We use the Adam optimizer for training, and during training, we use an initial learning rate of $1.6 \times 10^{-3}$. A warm-up strategy is employed, where the Gaussians are optimized for 1,000 iterations, followed by the optimization of the entire framework for an additional 3,000 iterations. 
The frequency of pruning and densification in point clouds depends on depth types and varies across different tasks. All experiments were conducted using an NVIDIA RTX 4090 GPU.

% \vspace{-1em}

\begin{figure*}[t]
	\centering
	\includegraphics[width = 1\linewidth]{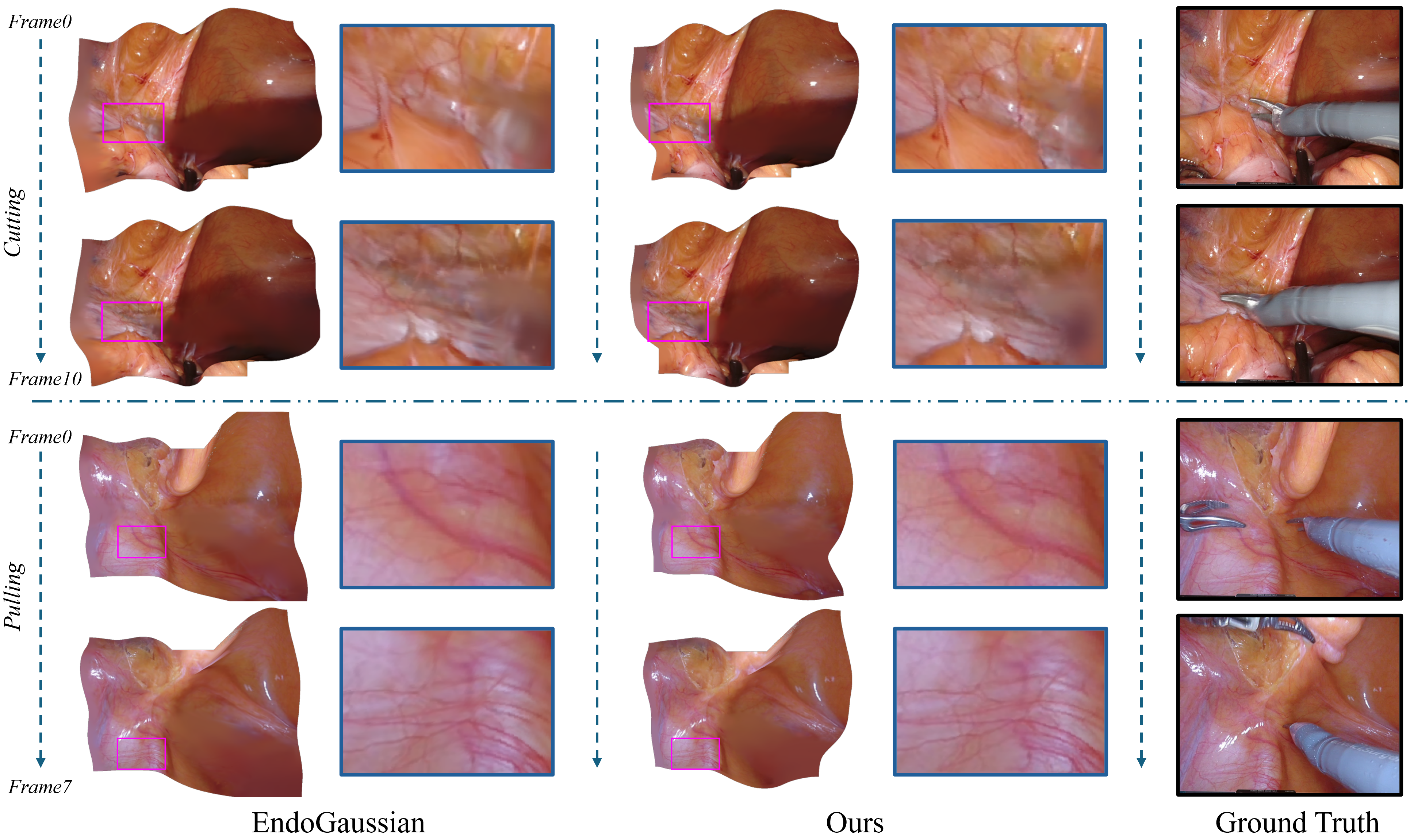}
        \caption{The qualitative result comparison between EndoGaussian and our proposed SAGS.}
\label{fig:qualitative_comparison}
\end{figure*}

\subsection{Comparison with the State-of-the-Art Methods}

We comprehensively compared our proposed SAGS framework with several SOTA methods across two benchmark datasets: EndoNeRF~\cite{endonerf} and SCARED~\cite{scared}, under both binocular and monocular depth settings. 

On the \textbf{EndoNeRF dataset}, SAGS achieves top performance in both binocular and monocular configurations. In the binocular setting, SAGS attains a PSNR of \textbf{39.164}, an SSIM of \textbf{0.970}, and an LPIPS of \textbf{0.025}, surpassing all previous approaches, including EndoGaussian (PSNR: 38.008, SSIM: 0.962, LPIPS: 0.048). In the monocular setup, SAGS also outperforms others with a PSNR of \textbf{37.711}, an SSIM of \textbf{0.962}, and an LPIPS of \textbf{0.043}, consistently achieving the best results across all metrics. These improvements reflect the model's robustness in reconstructing fine-grained structural and appearance details even under sparse input constraints.

On the \textbf{SCARED dataset}, which includes challenging scenarios with diverse lighting conditions and significant deformation, SAGS maintains consistent superiority. As reported in Table~\ref{comparison}, our method outperforms EndoNeRF, EndoSurf, and EndoGaussian across five different SCARED sequences. SAGS achieves the highest PSNR in four out of five sub-datasets (e.g., 33.53 on d2k1, 30.23 on d1k1), and its average performance across all sequences reaches a PSNR of \textbf{30.56}, SSIM of \textbf{0.866}, and LPIPS of \textbf{0.161}. This highlights the framework’s effectiveness in generalizing to highly dynamic and diverse surgical scenes beyond a single dataset.

Across datasets, SAGS consistently perform better than previous methods in terms of both pixel-level accuracy (PSNR/SSIM) and perceptual quality (LPIPS). Notably, even compared with recent high-performing models like EndoGS and LerPlane-32k, SAGS achieves better structural similarity and lower perceptual error, demonstrating its capacity to preserve geometric continuity and texture integrity under both sparse and dense depth supervision. 

Qualitative comparisons further confirm the effectiveness of our framework, as shown in Figure~\ref{detail}. Figure~\ref{detail} presents visual results for four challenging frames from the EndoNeRF dataset with binocular depth that involve strong specular highlights, rapid non-rigid tissue motion, and instrument-induced occlusions. In each case, the top row shows the full rendering, whereas the bottom row enlarges the red region of interest to reveal fine-grained differences. Methods adapted from NeRF (\emph{EndoNeRF} and \emph{EndoSurf}) suffer from noticeable blur and ringing around specular highlights, and the high-frequency vascular textures on the peritoneum become smeared once the camera viewpoint changes. The two existing Gaussian-splatting baselines (\emph{EndoGaussian} and \emph{EndoGS}) improve sharpness but still exhibit aliasing along instrument edges and faint contours on dynamically deforming tissue; in addition, colour consistency across adjacent frames is occasionally lost, producing flicker artifacts.

By contrast, the proposed \textbf{SAGS} reconstruction is visually closest to the reference video. Vessel bifurcations and subtle surface folds remain crisp, specular reflections are neither over-sharpened nor haloed, and instrument boundaries appear well-defined without stair-step artefacts. The alias-free rasteriser suppresses moire patterns that are visible in the EndoGaussian results (see third frame), while the self-adaptive deformation decoder prevents the texture tearing seen in EndoGS when the grasper lifts tissue (row 3). Across all test frames, SAGS delivers more coherent shading, fewer high-frequency artefacts, and superior geometric integrity, qualitatively corroborating the quantitative gains reported in Tables~\ref{comparison} and~\ref{comparison2}.

Figure~\ref{fig:qualitative_comparison} compares the reconstructed 3D meshes of our SAGS pipeline with the most closely related baseline, EndoGaussian. Across both sequences, the EndoGaussian reconstructions exhibit noticeable texture drift: vascular patterns become blurred, and high-frequency highlights bleed across neighbouring. These artefacts are particularly evident in the “Cutting’’ sequence at Frames 0 and 10, where the specular streak along the liver surface spreads beyond its anatomical boundary. By contrast, SAGS preserves crisp vessel bifurcations and maintains a stable highlight footprint, indicating that the alias-free rasteriser successfully suppresses high-frequency noise. Geometric fidelity also improves: in the “Pulling’’ sequence, the surface around the grasper tip flattens in the EndoGaussian model, whereas our deformation-adaptive decoder reconstructs the local indentation, matching the ground-truth depth cue.

\begin{table}[t]
\centering
\caption{Ablation study evaluating the efficacy of each proposed module on the EndoNeRF-Pulling~\cite{endonerf} dataset.}
\label{tab:ablation_study}
\begin{tabular}{lccc}
\toprule
\textbf{Ablation Study} & \textbf{PSNR↑} & \textbf{SSIM↑} & \textbf{LPIPS↓} \\
\midrule
Baseline               & 37.18          & 0.9577         & 0.0632          \\
w/o Alias-Free          & 37.33          & 0.9578         & 0.0627          \\
w/o SAD              & 38.08          & 0.9629         & 0.0434          \\
\textbf{SAGS (Full)}   & \textbf{38.34} & \textbf{0.9642} & \textbf{0.0326} \\
\bottomrule
\end{tabular}
% \vspace{-2em}
\end{table}

\subsection{Ablation Study}

To measure the contributions of each proposed component in the SAGS framework, \textit{i.e.}, the 3D smoothing filters and the self-adaptive deformation decoder (dubbed as ``SAD") module, we designed an ablation study on the EndoNeRF-Pulling dataset. 
Table~\ref{tab:ablation_study} presents the ablation study results on the EndoNeRF-Pulling dataset, evaluated using PSNR, SSIM, and LPIPS metrics. 
% Starting from the baseline model, adding the 3D smoothing filter slightly improves PSNR and LPIPS, showing its effectiveness in artifact reduction. Replacing the proposed SAD module with a single MLP
% leads to a noticeable drop in performance, highlighting its importance for handling deformations. The full SAGS model achieves the best results, demonstrating the synergy of all components.

\subsubsection{Effectiveness of Alias-Free Processing}
Endoscopic videos contain strong specular highlights, sharp tissue boundaries, and rapid motion, all of which amplify high-frequency content and make dynamic reconstructions especially susceptible to ringing, Moire patterns, and shimmer across frames.  Hence, suppressing aliasing is critical for delivering clinically reliable 3D visualisation. We first evaluate the effectiveness of the Alias-Free processing module. As illustrated in Table~\ref{tab:ablation_study}, adding Alias-Free processing to the baseline model results in improvements in both PSNR (from 37.18 to 37.33) and LPIPS (from 0.0632 to 0.0627). These improvements, although subtle, demonstrate that the Alias-Free module effectively mitigates high-frequency noise and aliasing artifacts, contributing to clearer and smoother visual reconstructions. This validates the importance of the Alias-Free processing in enhancing visual quality and reducing perceptual artifacts in dynamic surgical scene reconstruction.

\subsubsection{Effectiveness of SAD Module}

Modelling surgical tissue motion is particularly challenging: organs undergo centimetre-scale, non-linear deformations, and the interaction with graspers or scissors can introduce abrupt, topology-changing displacements.  
Static or fixed-weight networks often fail to track these rapid, heterogeneous motions, leading to texture drift and geometric distortion over time.
To quantify the benefit of our \emph{Self-Adaptive Weighted Deformation (SAD)} module—equipped with dynamic multi-head attention and learnable per-head weights—we replace it with a single-layer MLP of comparable parameter count.  
Removing SAD causes a consistent drop across all metrics: PSNR falls from 38.34 to 38.08, SSIM from 0.9642 to 0.9629, while LPIPS rises from 0.0326 to 0.0434.  
These degradations confirm that the SAD module is crucial for capturing the intricate spatial–temporal variations of soft tissue, enabling perceptually faithful and geometrically accurate reconstructions in dynamic endoscopic scenes.

\subsubsection{Effectiveness of Combined Modules}
Lastly, we examine the combined impact of integrating both the Alias-Free processing and the SAD module within the full SAGS framework. The complete model achieves the best results, achieving a PSNR of 38.34, SSIM of 0.9642, and LPIPS of 0.0326. This substantial performance gain illustrates a clear synergistic effect, where the complementary functions of artifact suppression by the Alias-Free module and dynamic modeling capabilities by the SAD module effectively enhance the overall reconstruction quality. Thus, this final evaluation confirms the integral roles and synergistic relationship of these two key components, jointly addressing challenges posed by dynamic endoscopic scenes.

\section{Conclusion}
In this work, we introduced SAGS, a novel self-adaptive alias-free Gaussian splatting framework for dynamic endoscopic scene reconstruction. Leveraging a dynamically weighted deformation decoder with multi-head attention and advanced alias-free processing through 3D smoothing and 2D Mip filters, SAGS effectively reduces artifacts and captures fine-grained tissue details under complex deformations. Comprehensive evaluations on EndoNeRF~\cite{endonerf} and SCARED~\cite{scared} datasets demonstrated that our SAGS outperforms state-of-the-art methods in terms of PSNR, SSIM, and LPIPS, highlighting its ability to preserve geometric and texture fidelity. Beyond quantitative improvements, qualitative results further validated the superiority of SAGS in reconstructing sharp details, mitigating aliasing, and maintaining temporal consistency in challenging scenarios involving dynamic tissue deformations and surgical tool interactions. These results emphasize the potential of our method for robotic-assisted surgery, where precise 3D modeling is essential for navigation and intervention planning.

\bibliographystyle{elsarticle-harv}
\bibliography{sample-base}

\end{document}